\documentclass[conference]{IEEEtran}
\IEEEoverridecommandlockouts
\usepackage{cite}
\usepackage{amsmath,amssymb,amsfonts}
\usepackage{algorithmic}
\usepackage{graphicx}
\usepackage{textcomp}
\usepackage{xcolor}
\def\BibTeX{{\rm B\kern-.05em{\sc i\kern-.025em b}\kern-.08em
    T\kern-.1667em\lower.7ex\hbox{E}\kern-.125emX}}

\usepackage[utf8]{inputenc}
\usepackage{blindtext}
\usepackage{multicol}

\usepackage{times}
\usepackage{epsfig}
\usepackage{graphicx}
\usepackage{amsmath}
\usepackage{amssymb}

\usepackage{subcaption}
\usepackage{mathrsfs}
\usepackage{booktabs}
\usepackage{multirow}
\usepackage{array}
\usepackage{comment}
\usepackage{caption}
\usepackage{rotate}
\usepackage{multirow}
\usepackage{subcaption}
\usepackage{hyperref}

\usepackage[letterpaper,top=2cm,bottom=2cm,left=3cm,right=3cm,marginparwidth=1.75cm]{geometry}
\usepackage{xcolor}

\graphicspath{ {./images/} }

\begin{document}

\title{High Efficiency Pedestrian Crossing Prediction}

\author{\IEEEauthorblockN{ Zhuoran Zeng}
\IEEEauthorblockA{\textit{New York University} \\
\textit{Dept. of Computer Science}\\
zz2017@nyu.edu}
}

\maketitle




\newcommand{\zw}{\color{blue}}

\begin{abstract}
Predicting pedestrian crossing intention is an indispensable aspect of deploying advanced driving systems (ADS) or advanced driver-assistance systems (ADAS) to real life. State-of-the-art methods in predicting pedestrian crossing intention often rely on multiple streams of information as inputs, each of which requires massive computational resources and heavy network architectures to generate. However, such reliance limits the practical application of the systems. In this paper, driven the the real-world demands of pedestrian crossing intention prediction models with both high efficiency and accuracy, we introduce a network with only frames of pedestrians as the input. 
Every component in the introduced network is driven by the goal of light weight. Specifically, we reduce the multi-source input dependency and employ light neural networks that are tailored for mobile devices. These smaller neural networks can fit into computer memory and can be transmitted over a computer network more easily, thus making them more suitable for real-life deployment and real-time prediction. To compensate the removal of the multi-source input, we enhance the network effectiveness by adopting a multi-task learning training, named "side task learning", to include multiple auxiliary tasks to jointly learn the feature extractor for improved robustness. Each head handles a specific task that potentially shares knowledge with other heads. In the meantime, the feature extractor is shared across all tasks to ensure the sharing of basic knowledge across all layers. The light weight but high efficiency characteristics of our model endow it the potential of being deployed on vehicle-based systems. Experiments validate that our model consistently delivers outstanding performances.








\end{abstract}
\section{Introduction}
Recent years has witnessed an increase in pedestrian fatalities. While definite reasons of rising pedestrian fatalities remain unclear, they can roughly be attributed to distracted driving, speeding, and reckless driving. With the development of advanced visual sensors such as front-facing camera and even 360 degree videos , it becomes possible for vehicles rather than drivers to monitor pedestrians and road situations. More than ever, empowering vehicles to recognize and predict pedestrian behaviors becomes unprecedentedly important and indispensable.

Pedestrian crossing prediction has been explored for years in academia. Early works usually utilize a single frame as input to a convolutional neural network (CNN) model to generate predictions \cite{JAADDataset}. However, this approach ignores the temporal aspects of videos and contextual data. Later, with the maturity of recurrent neural networks (RNNs), pedestrian crossing intention was improved by considering both spatial and temporal information as well as including contextual information \cite{ContextPIP} \cite{2DPosePCI}, e.g., pedestrian bounding box, pose estimation, behaviors, appearances, vehicle information, road situations and etc. More recently, research has been focusing on different ways of fusing multi-stream features\cite{PCI}. However, the use of multi-stream features as input increases latency and requires massive additional computational resources, thus making the models almost impossible to deploy to real-life driving systems. 

In this work, driven by the principle of light weight and high efficiency, we propose a new neural network based model. The newly-proposed model achieves both high efficiency and effectiveness from the following two perspectives. First, we reduce the dependency on different input channels and employ smaller neural networks, such as SqueezeNet\cite{SqueezeNet} and MobileNets\cite{MobileNets}, as the main feature extractor. These neural networks have much fewer parameters and thus can fit into mobile devices more easily. Moreover, the smaller size enables our model to be transmitted over computer networks at ease, with shorter transmitting time and less computational resources.

In addition, inspired by the literature of multi-task learning\cite{MTL}, we adopt an approach named ``side-task learning" to include multiple auxiliary task-specific heads, each of which handles a specific task that potentially shares knowledge with the main intention prediction head. The motivation of such deign is to reintroduce crucial information such as segmentation and pedestrian poses excluded from input sources without explicitly adopting such information as inputs. The feature extractor is shared across all tasks to ensure the sharing of basic knowledge across all layers. On top of the feature extractor, we devise two tasks, one for predicting pedestrian crossing intention and the other for estimating pedestrian poses. We hope that the model can better predict crossing intention with the knowledge of pedestrian poses. 

In summary, our contributions are as follows:
\begin{itemize}
    \item Our newly-proposed model is light-weight in that we reduce dependency on multi-stream of input sources and employ smaller neural networks as the main feature extractor. The light weight of the model endows it the potential of being deployed to real-life driving systems. 
    \item Our newly-proposed model utilizes ``side-task learning", a variation of multi-task learning. It includes multiple task-specific heads, such as one for predicting pedestrian crossing intention and one for estimating pedestrian poses, and share knowledge across different layers. In this way, we facilitate knowledge transfer among different tasks and thus improving learning efficiency and prediction accuracy.
    \item We validate the performance on real-world data, and the model consistently achieves state-of-the-art performance. 
\end{itemize}

\section{Related Work}

The study of vision-based pedestrian crossing prediction traces back to Caltech Pedestrian Detection Benchmark \cite{caltech}. Caltech dataset collected videos taken from a vehicle driving through regular traffic in an urban environment and provides bounding box for pedestrians in each frame. However, it does not provide annotations for pedestrian behaviors (such as standing, looking, and walking), pedestrian appearances (such as male or female, glasses or no glasses), or contextual information (such as speed, stop signs). This gap is later filled by the JAAD dataset\cite{JAADDataset}, which offers explicit pedestrian annotations and contextual information. With the introduction of JAAD dataset, a baseline method for scene analysis is also proposed. The baseline method\cite{JAADDataset} takes into consideration both static environmental context and behavior of the pedestrians and utilizes AlexNet together with a linear support vector machine (SVM) to predict crossing or not crossing event.

Many tasks are involved in predicting pedestrian crossing intentions. In this section, we will review the tasks in predicting pedestrian crossing intention as well as the techniques we utilize in our model.

\subsection{Autonomous Driving with Computer Vision}

With the raise of deep learning, we witness a wide range of computer vision applications on autonomous driving. 

Object detection is adapted to pedestrian detection, vehicle detection, lane detection, and etc. Despite having different tasks, these systems often share some common object detectors such as Faster R-CNN \cite{fasterRCNN} or YOLO \cite{YOLO}. Faster R-CNN belongs to the two-stage object detector category, where the first stage is to generate regional proposals, potential regions that may contain an object, and the second stage is to classify and localize the previously proposed regions. YOLO belongs to single-stage object detector, where it uses predefined boxes/keypoints of various scale and aspect ratio to classify and localize objects in one shot. Many detection tasks related to autonomous driving are built on top of these backbone architectures. 

Semantic segmentation is another important task that facilitates autonomous driving. The goal of semantic segmentation is to densely assign class label to each pixel in the input image for precisely understanding of the scene. Therefore, in the past decades, a significant amount of work has been dedicated to treating semantic segmentation as image classification at pixel level \cite{long2015fully} \cite{paszke2016enet}. But this conventional method doesn't take into consideration of the importance of different pixels. This gap is later filled by importance-aware methods \cite{liu2020importanceaware} which argue that the distinction between object/pixel importance need to be taken under consideration. Having important-aware segmentation methods better emphasize the objects we want to study in autonomous driving such as pedestrians or vehicles. 

Apart from RGB image based algorithm, methods based on other sensors like LiDAR\cite{lidaroverview} and radar\cite{radar_overview} are also widely adopted for 3D objection detection for improved robustness against lighting. LiDAR stands for "Light Detection and Ranging", a method of measuring distance by shooting lasers and detecting how much time they take to return to decide distance of objects. Radar shares similar ideas with LiDAR, except it uses radio waves to decide such distance. Compared to traditional vision-based sensors, LiDAR and radar sensors can generate 3D representation of the surroundings and have the advantages of higher detection accuracy and higher speed \cite{lidar_on_ped} \cite{radar}. 

\subsection{Multi-Task Learning}

In general, when we want to solve several tasks at the same time, we may attempt to train several models, each of which solves a task independently. However, this approach ignores the information that may be helpful to our learning, especially when tasks are related. This is where Multi-Task Learning (MTL) comes in. Multi-Task Learning (MTL)\cite{MTL} is an approach that solves multiple learning tasks simultaneously and shares knowledge among different tasks. 

There are two MTL methods for deep learning: hard and soft parameter sharing. Hard parameter sharing\cite{MTLhard} is the most commonly used approach to MTL in neural networks. It is generally applied by sharing the hidden layers between all tasks, while keeping several task-specific output layers. On the other hand, in soft parameter sharing, each task has its own model with its own parameters. The distance between the parameters of the model is then regularized in order to encourage the parameters to be similar \cite{MTLsoft1} \cite{MTLsoft2}. Compared to training the models separately, multi-task learning reduced the risk of overfitting \cite{MTL_overfitting} and improved generalization by making use of domain knowledge learned from other related tasks; what we learned from one task helps other tasks learn better. 

\subsection{Efficient Neural Networks and Techniques}

In our experiment, we utilize efficient neural networks such as SqueezeNet\cite{SqueezeNet} and MobileNet\cite{MobileNets} as the main feature extractor. SqueezeNet reduces the number of parameters while maintaining the performance through clever design strategies: replace 3×3 filters with 1×1 filters, decreasing the number of input channels to 3x3 filters, and placing downsampling layers later in the network. MobileNets, on the other hand, adopts a different architecture design. It factorizes a standard convolution into a depthwise convolution and a 1x1 pointwise convolution. Compared to a standard convolution, which uses kernels on all input channels and combines them in one step, the depthwise convolution separates the kernel filter for each input channel and uses pointwise convolution to combine inputs. This separation of filtering and combining of features reduces the computation cost and model size.

Other than efficient network design, various compression techniques, such as pruning \cite{pruning1,pruning2,pruning3,pruning4}, \\ quantization\cite{quan1,quan2}, also help to reduce model size and improve the efficiency of neural networks. Pruning often refers to removing weights that are already close to zero, since their effect to the neural networks is almost negligible. Quantization is a technique to reduce the number of bits needed to store each weight in the Neural Network through weight sharing. It is done by mapping floats to integers and forming a ``Codebook" to map the original weight to the quantized weight. Deep Compression combines both pruning and quantization, along with Huffman encoding which is used to reduce the amount of bits needed to represent the weights in the quantized ``Codebook".

\newcommand{\batch}{\mathbf{X}}
\newcommand{\img}{\mathbf{x}}
\newcommand{\model}{\boldsymbol{\Phi}}
\newcommand{\FE}{\mathbf{E}}
\newcommand{\posehead}{\mathbf{H_p}}
\newcommand{\crosshead}{\mathbf{H_c}}

\section{Method}

\paragraph{Problem Formulation}
Vision-based pedestrian crossing intention prediction is formulated as follows: given a sequence of video frames from the front camera a vehicle, a model estimates the probability of the target pedestrian p’s action $A^{t+n}_p \in \{0,1\} $  of crossing the road, where $t$ is the specific time of the last observed frame and $n$ is the number of frames from the last observed frame to the crossing / not crossing (C/NC) event.

In the proposed model, features such as pedestrian bounding box and pedestrian poses are provided. Pedestrian bounding box is used to crop image around the pedestrians, and pedestrian poses serve as ground-truth labels to facilitate multi-task learning. Input to this model is a sequence of video frames where only one pedestrian is included. The model takes in this sequence as input and has two heads in the output layer, one for pedestrian crossing intention prediction and the other for pedestrian pose prediction. Thus, our model can be represented as this: 

\begin{itemize}
    \item Model $\boldsymbol{\Phi}$ is composed of a feature extractor $\FE$ and two heads $\crosshead$ and $\posehead$, parametrized by $\theta_c$ and $\theta_p$ respectively.
    
    \item Video clip $\img \in \mathbb{R}^{l \times c \times h \times w}$, where $l$ is the length of the video clips, $c$ is the input image channel, $h$ is the height of the input image, and $w$ is the width of the input image. 

    \item Prediction of crossing $\mathbf{\hat{y}_\text{cross}} = \crosshead(\FE(\img); \theta_c)$ will be compared with ground-truth label $\mathbf{y_\text{cross}}$. 
    
    \item Pose prediction $\mathbf{\hat{y}_\text{pose}} = \posehead(\FE(\img); \theta_p)$ will be compared with ground-truth label $\mathbf{y_{\text{pose}}} \in \mathbb{R}^{k \times 2}$, where $k$ is the number of keypoints. 
\end{itemize}

\subsection{Feature Extractor}





We utilize SqueezeNet, a light-weight convolutional neural network, as the main feature extractor. Input to the feature extractor is a batch of video clips $\batch \in \mathbb{R}^{n \times l \times c \times h \times w}$, where $n$ is the number of videos in a batch, $l$ is the length of the video clips, $c$ is the input image channel, $h$ is the height of the input image, and $w$ is the width of the input image. This batch is then reshaped to size $[n \times l, c, h, w]$ to feed into SqueezeNet. After going through SqueezeNet, the input batch will have the feature dimension of $[n \times l, 512]$.

\subsection{Crossing Prediction} 
To predict pedestrian crossing intention, we use gated recurrent unit (GRU) \cite{GRU} combined with a linear layer in the end. The reason for choosing GRU is that it is more computationally efficient than its counterpart LSTM \cite{LSTM}, which is older, and its architecture is relatively simple. 

\label{eq:GRU} 
\begin{align*}
& z_t = \sigma_g(W_zx_t + U_zh_{t-1} + b_z) \\
& r_t = \sigma_g(W_rx_t + U_rh_{t-1} + b_r) \\
& \hat{h}_t = \phi_h(W_hx_t + U_h(r_t \odot h_{t-1}) + b_h) \\ 
& h_t = (1-z_t) \odot h_{t-1} + z_t \odot \hat{h}_t
\end{align*}

where $x_t$ is the input vector, $h_t$ is the output vector, $\hat{h}_t$ is the candidate activation vector, $z_t$ is the update gate vector, $r_t$ is the reset gate vector, $W$, $U$, and $b$ are parameter matrices and vector. $\sigma_g$ is a sigmoid function and $\phi_h$ is hyperbolic tangent originally.

Feature tensor is of size $[n \times l, 512]$ after going through the feature extractor. This tensor is then reshape into size $[l, n, 512]$ to fit the input size of GRU. The GRU we applied has hidden size of 512, which results in the output tensor of size $[l, n, 512]$. We will only utilize the prediction from the last frame, which reduces the tensor size to $[n, 512]$. It is then passed into a linear layer with output channel of size 2. Thus, crossing prediction tensor eventually ends with size of $[n, 2]$.

\subsection{Auxiliary Supervision}

In order to learn a more robust feature extractor and stimulated by the literature of multi-task learning, we introduce ``side-task learning". Specifically, we impose auxiliary predictions heads $\{h_1, h_2, \dots, h_n\}$, each of which is in charge of one particular tasks that we believe has shared knowledge with crossing prediction and has special tailored architecture.

\paragraph{Pose Prediction Module}
Pose prediction module contains two linear layers, one has input channel 512 and output channel 512, and the other has input channel 512 and output channel 36, which is the number of keypoints times 2 for x and y coordinates. Between these two linear layers, the model also encompasses a batch normalization layer and a ReLU nonlinearity to normalize and transform data. We also apply dropout layer of probability $0.5$ to achieve better training performance. A sigmoid nonlinearity is then appended at the end to transform data between 0 and 1. The goal is to return the ratios of the actual $x$, $y$ coordinates of predicted poses to the size of the image.

\paragraph{Speed Prediction Module} 
We also include a speed head along with the pose head to do side-task learning, hoping that the information of vehicle speed facilitates the model's learning. For the speed head layer, we adopt a similar set-up as for pose head, which mostly include linear layers and data normalization/transformation layers. 

\section{Experiments}

\subsection{Set-up}
\paragraph{Dataset}
In this work, we adopt Joint Attention in Autonomous Driving (JAAD) Dataset. First proposed in \cite{JAADDataset}, JAAD dataset focuses on pedestrian and driver behaviors at the point of crossing and factors that influence them. To achieve this end, JAAD dataset provides a richly annotated collection of 346 short video clips (5-10 sec long) extracted from over 240 hours of driving footage. Its annotations include spatial annotations (bounding box), pedestrian behavioral annotations (walking, standing, looking, etc), pedestrian attributes (age, gender, clothing, accessories, etc), and contextual annotations (elements of infrastructure, weather, etc). These annotations are provided for each pedestrian per frame. 




\begin{figure}
\centering
\begin{subfigure}{0.5\linewidth}
  \includegraphics[width=\textwidth]{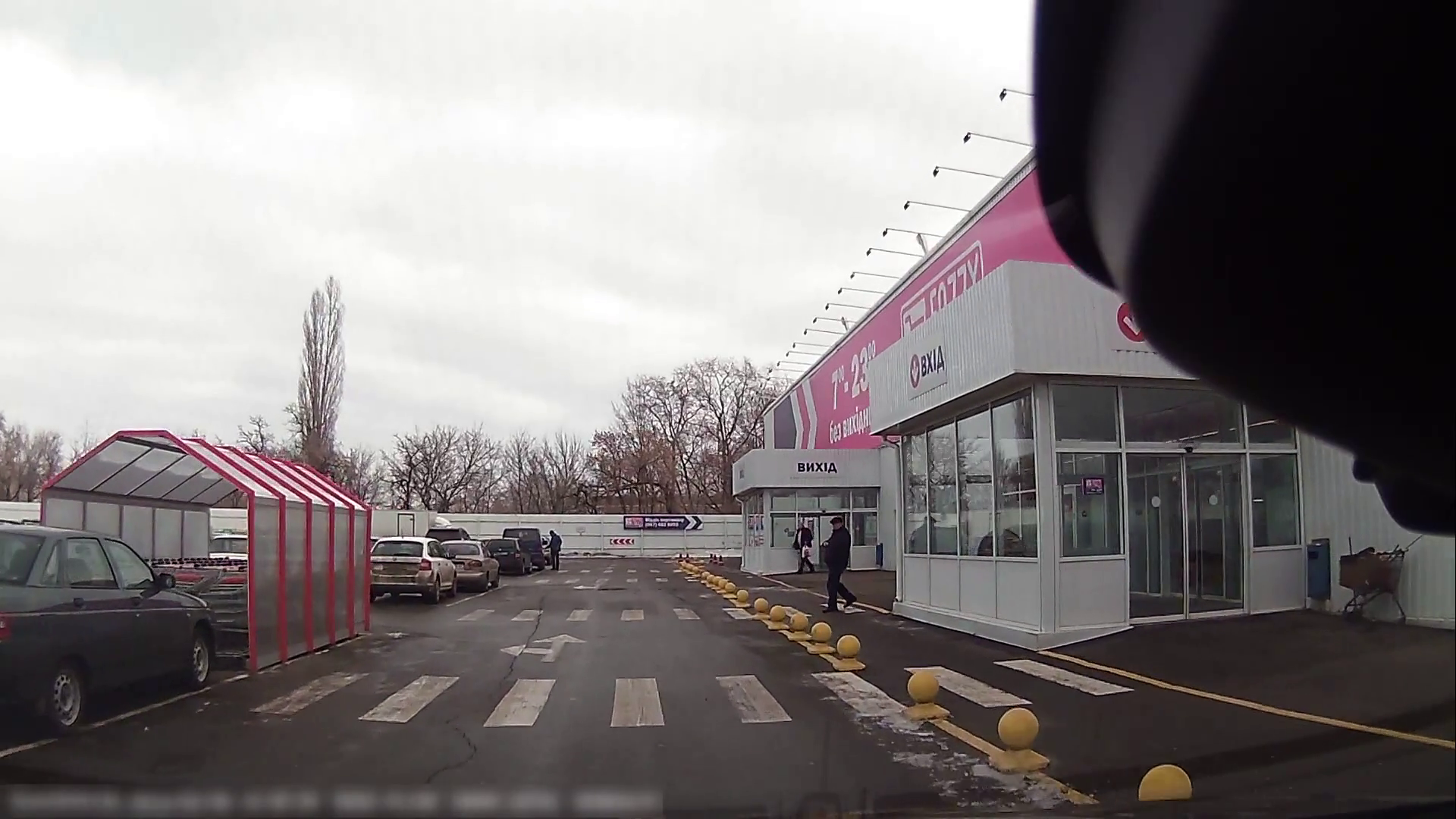}
  \label{fig:sub1}
\end{subfigure}%
\begin{subfigure}{0.5\linewidth}
  \includegraphics[width=\textwidth]{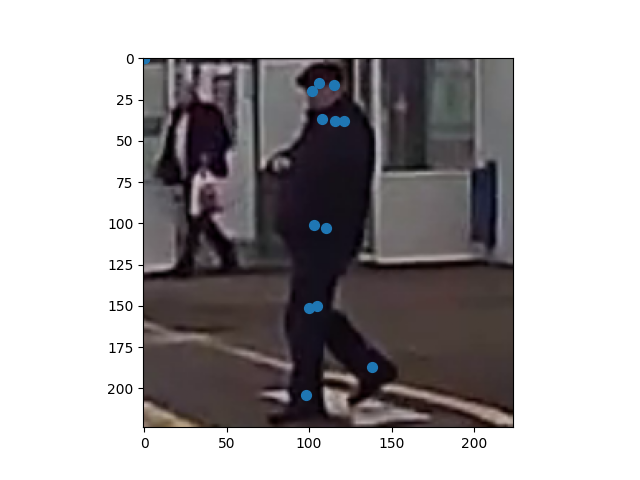}
  \label{fig:sub2}
\end{subfigure}

\begin{subfigure}{0.5\linewidth}
  \includegraphics[width=\textwidth]{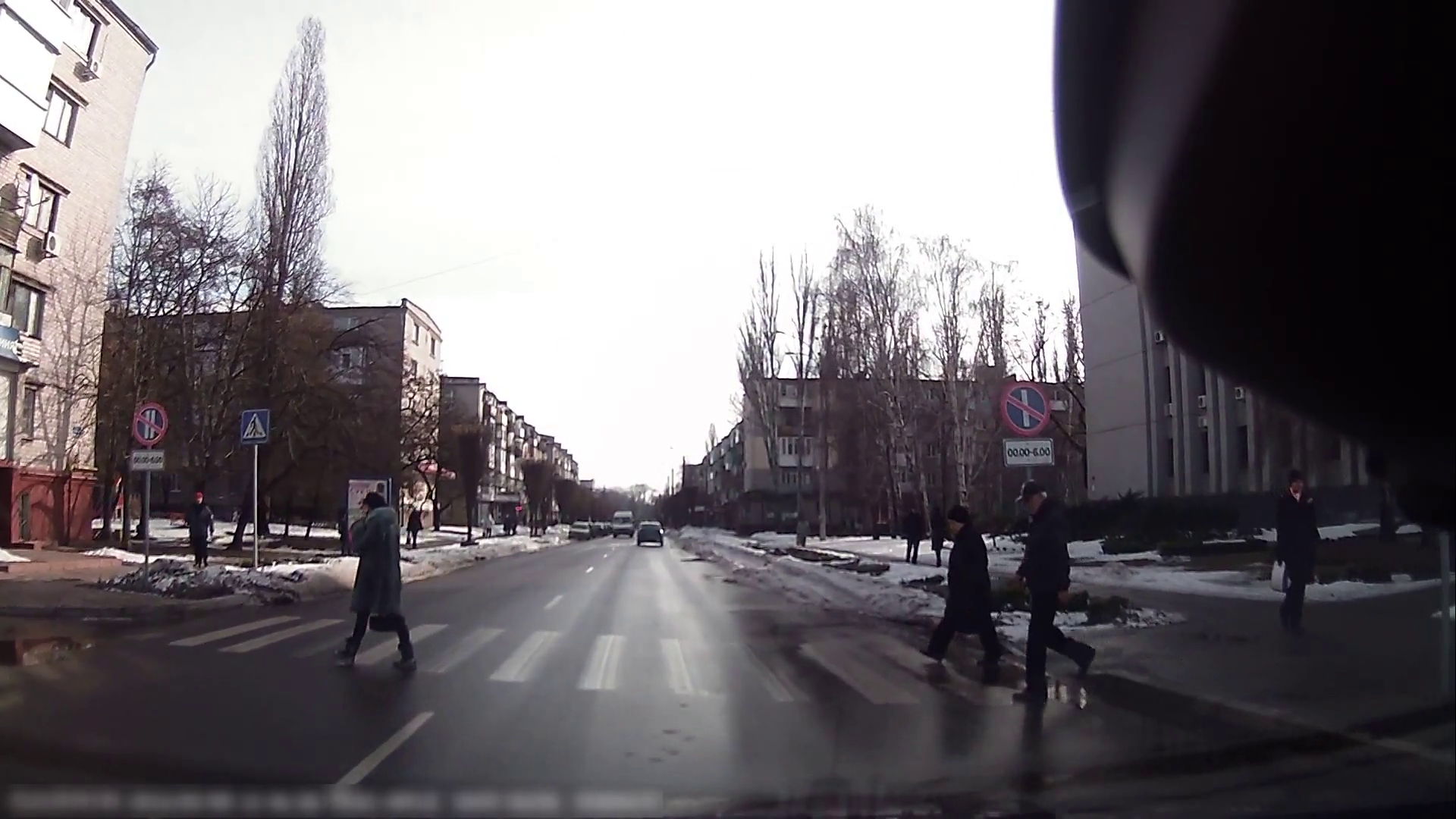}
  \label{fig:sub3}
\end{subfigure}%
\begin{subfigure}{0.5\linewidth}
  \includegraphics[width=\textwidth]{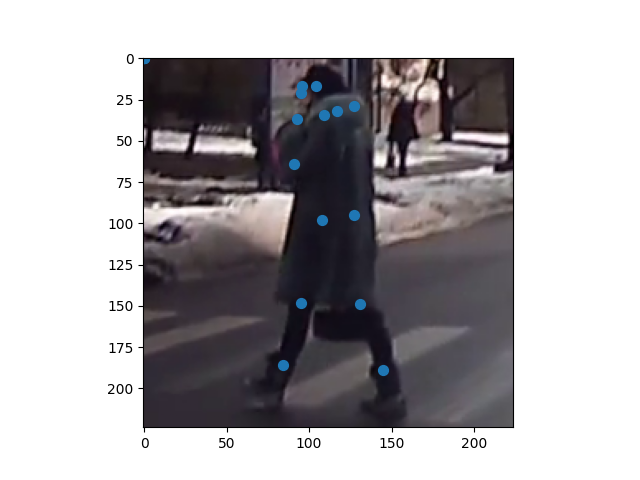}
  \label{fig:sub4}
\end{subfigure}

\caption{Dataset example. On the left is original frames from videos; 
on the right is the images fed into the model, cropped by bounding box and annotated with pose information}
\label{fig:test}
\end{figure}

We used JAAD dataset for training and testing our proposed model. JAAD dataset provides two subsets, JAAD behavioral data ($\text{JAAD}_\text{beh}$) and JAAD all data ($\text{JAAD}_\text{all}$). $\text{JAAD}_\text{beh}$ contains pedestrians who are crossing (495) or are about to cross (191). $\text{JAAD}_\text{all}$ has additional pedestrians (2100) with non-crossing actions.

We process our training and testing dataset in the following way. Video clips are split into frames initially. Then video sequence are aggregated based on pedestrian ids, ie. each pedestrian has a video sequence of their own. These video sequences are further expanded by taking every 16-frame into a new video sequence. The resulting collection of video sequences constructs our training or testing set. 

\paragraph{Metric}
For the main crossing prediction task, we utilize accuracy as the main indicator of the effectiveness of our model and adopt cross entropy loss to measure losses. For the auxiliary pose and speed prediction task, we adopt binary cross entropy loss. The overall loss function is expressed as follows: 
\begin{center}
    $\text{loss} = \text{loss}_\text{cross} + \lambda \times \text{loss}_\text{pose} + \lambda \times \text{loss}_\text{speed}$
\end{center}

where $\lambda$ represents the weight we want to count in the loss of predicting poses. 

\paragraph{Detail}


On hardware level, We use AMD Ryzen 5 3600 as CPU and Nvidia GeForce RTX 3070 as GPU. Having nearly 6000 cores, GeForce RTX 3070 makes a good candidate for processing multiple computations simultaneously, thus facilitating us in reaching a decent accuracy. In considering library, we choose PyTorch over Tensorflow in that PyTorch is more modular, which makes it easier to break down the phase of preprocessing dataset and the phase of the acutal training and testing. 

On implementation level, we adopt Adam optimizer with default learning rate of 1e-2 and default weight decay of 1e-5. We also use MultiStepLR as learning rate scheduler with milestones 50 and 75 and gamma of 0.1. In calculating the loss of crossing prediction, we pass in weight of $[1760.0/2134.0, 1-1760.0/2134.0]$ since JAAD dataset has an unbalanced number of crossing (1760) and non-crossing (374) events. 

\subsection{Ablation Study}

\paragraph{Hyperparameter $\lambda$}
$\lambda$ appears as a multiplier of the pose loss and the speed loss in the total loss function and is used to balance the loss terms. We experiments with different values of $\lambda$ to see how including side task learning facilitate our main task prediction. The results show that, between $[0, 0.1]$, when $\lambda$ tends to $0.1$, the accuracy of the main task drops. One sweet spot is around $\lambda = 0.01$, during which the accuracy achieves its max value $(84 \% )$. More comparisons are in Table \ref{tab:acc}.

\paragraph{Training Dataset} 
As mentioned before, JAAD dataset provides JAAD behavioral data ($\text{JAAD}_\text{beh}$) and JAAD all data ($\text{JAAD}_\text{all}$). The difference is that $\text{JAAD}_\text{all}$ has some additional non-crossing instances. We adjusted the weight accordingly to see how these two datasets perform. For $\text{JAAD}_\text{beh}$, we use the weight $[1760.0/2134.0, 1-1760.0/2134.0]$ since it has 1760 crossing events and 374 non-crossing events. When training with $\text{JAAD}_\text{all}$, we instead adjust the weight to $[1760.0/8613.0, 1-1760.0/8613.0]$ since $\text{JAAD}_\text{all}$ has 1760 crossing events and 6853 non-crossing events.

Using $\text{JAAD}_\text{all}$ as training dataset achieves better performance than $\text{JAAD}_\text{beh}$. The results show that training with $\text{JAAD}_\text{all}$ achieves accuracy around $20 \%$ more than training with $\text{JAAD}_\text{beh}$. Although $\text{JAAD}_\text{all}$ is more unbalanced than $\text{JAAD}_\text{beh}$, incorporating more samples with different contextual information help increase the accuracy. 

\paragraph{Performance Evaluation}
Besides accuracy, we also attempt other performance evaluation metrics, such as precision and receiver operating characteristics (ROC). The motivation behind this is, first, both $\text{JAAD}_\text{beh}$ and $\text{JAAD}_\text{all}$ dataset are very unbalanced. Suppose $90\%$ of the training data is negative instances, the model may learn to predict any unseen data as negative in order to achieve accuracy of around $90\%$. Second, under the context of predicting pedestrian crossing intention, we want to minimize false negative cases as much as possible and tolerate false positive cases to some extent. What this means is that, for those pedestrians who actually crosses, we want to predict them as crossing; for those that don't cross, we can tolerate them being classified as crossing. But we want to avoid or minimize the case where we predict a crossing pedestrian as non-crossing. Taking the above two motivations into consideration, we adopt precision and ROC as extra metrics to evaluate the performance of our model.  

\begin{figure}
    \centering
    \includegraphics[scale=0.4]{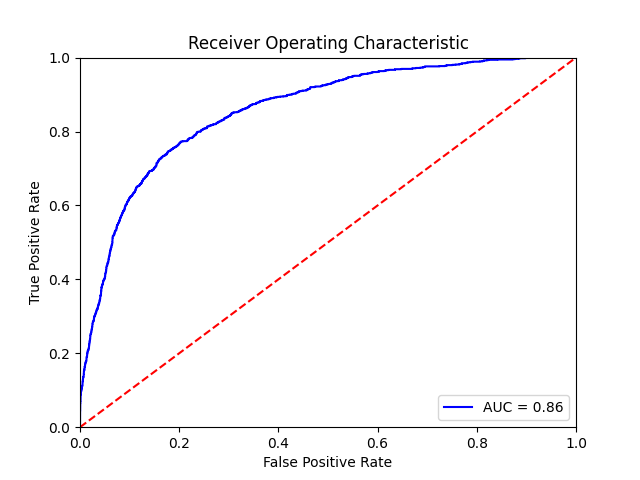}
    \caption{AUC-ROC curve generated from MobileNets on $\text{JAAD}_\text{all}$ dataset with $\lambda = 0.01$}
    \label{fig:roc}
\end{figure}

\subsection{Results}


The quantitative result is shown at Table~\ref{tab:acc} and  Table~\ref{tab:auc}, which reports the accuracy and auc-roc score of training with different backbone architectures and with different datasets respectively. We used MobileNets and SqueezeNet as backbone architecture and trained on both $\text{JAAD}_\text{beh}$ and $\text{JAAD}_\text{all}$ datasets. 

\begin{table}[h]
    \centering
    \caption{Accuracy with different backbone architectures and $\lambda$.}
    \label{tab:acc}
    \begin{tabular}{c|c|c c c}
        \toprule
         Backbone & Dataset & $\lambda = 0$ & $\lambda = 0.01 $ & $\lambda = 0.1$  \\
        \midrule
        MobileNet  & JAAD all & 81.33 & 84.04 & 82.19 \\
        SqueezeNet & JAAD all & 82.66 & 84.27 & 83.35 \\
        MobileNet  & JAAD beh & 58.75 & 57.89 & 60.09\\
        SqueezeNet & JAAD beh & 62.77 & 60.95 & 60.91\\
        \bottomrule
    \end{tabular}
\end{table}

The result in Table I shows that including side tasks facilitate the prediction of the main task and thus improving the overall performance. For $\text{JAAD}_\text{all}$ dataset and across both MobileNets and SqueezeNet architectures, when side tasks have weight $\lambda = 0.01$, we achieves an accuracy of around $84 \%$, the best accuracy result across all experiments. For MobileNets on $\text{JAAD}_\text{beh}$ dataset, $\lambda = 0.1$ reaches the best performance among other $\lambda$ configurations. However, numbers are less consistent for SqueezeNet where $\lambda = 0$ reaches the best performance. This means that auxiliary heads fail to provide useful information to the learning for SqueezeNet. 

\begin{table}[h]
    \centering
    \caption{AUC score with different backbone architectures and $\lambda$.}
    \label{tab:auc}
    \begin{tabular}{c|c|c c c}
        \toprule
         Backbone & Dataset & $\lambda = 0$ & $\lambda = 0.01 $ & $\lambda = 0.1$  \\
        \midrule
        MobileNet  & JAAD all & 0.83 & 0.85 & 0.84 \\
        SqueezeNet & JAAD all & 0.83 & 0.83 & 0.83 \\
        MobileNet  & JAAD beh & 0.55 & 0.53 & 0.55\\
        SqueezeNet & JAAD beh & 0.53 & 0.53 & 0.55\\
        \bottomrule
    \end{tabular}
\end{table}

In Table~\ref{tab:auc}, we show the AUC scores. As mentioned above, we want to minimize false negative cases as much as possible but at the same time tolerate false positive cases to some extent. AUC score is a good indicator for this purpose. It measure the ability of the classifier to distinguish between positive and negative cases. When training on $\text{JAAD}_\text{all}$, $\lambda = 0.01$ reaches the best AUC score overall. For MobileNets training on $\text{JAAD}_\text{beh}$, $\lambda = 0$ and $\lambda = 0.1$ has the same AUC score. For SqueezeNet, $\lambda = 0.1$ has the best AUC score. The results of AUC scores follow a similar pattern with the accuracy.

\section{Conclusion}


In this work, we proposed a new architectural design for vision-based pedestrian crossing prediction. Our design utilized light-weight models, such as MobileNets and SqueezeNet, to achieve resources efficiency. The adoption of side-task learning strategy enabled the model to learn as much as contextual information as possible. We ran experiments on both For $\text{JAAD}_\text{beh}$ and For $\text{JAAD}_\text{all}$ dataset and on the weight given to side tasks. Experiments show that our model achieves the state-of-the-art against baseline methods in the pedestrian crossing intention prediction benchmark with less complex architectural design and less resources. 

\medskip

\bibliographystyle{ieeetr}
\bibliography{main}

\end{document}